\documentclass[runningheads]{llncs}

\usepackage{eccv}

\usepackage{eccvabbrv}

\usepackage{graphicx}
\usepackage{booktabs}

\usepackage{multirow}
\usepackage{xcolor}

\usepackage[accsupp]{axessibility}  

\usepackage{hyperref}

\usepackage{orcidlink}

\begin{document}

\title{Beyond Training from Scratch:
Foundation Models for Data-Efficient and Generalizable Cardiac MRI Reconstruction} 


\author{Anam Hashmi\inst{1,2}\orcidlink{0009-0007-0887-246X} \and
Mayug Maniparambil\inst{2}\orcidlink{0000-0002-9976-1920} \and
Julia Dietlmeier\inst{2}\orcidlink{0000-0001-9980-0910} \and \\
Kathleen M. Curran\inst{3}\orcidlink{0000-0003-0095-9337} \and
Noel E. O’Connor\inst{2,4}\orcidlink{0000-0002-4033-9135}}

\authorrunning{A.~Hashmi et al.}

\institute{
Research Ireland Centre for Research Training in Machine Learning, Ireland \\
\and
Dublin City University, Ireland \\
\email{anam.hashmi2@mail.dcu.ie, mayug.maniparambil2@mail.dcu.ie, Julia.Dietlmeier@dcu.ie, Noel.OConnor@dcu.ie}
\and
University College Dublin, Ireland \\
\email{kathleen.curran@ucd.ie} \and
Rinn Artificial Intelligence, Ireland 
}

\maketitle

\begin{abstract}
  Cardiac magnetic resonance imaging reconstruction aims to recover high-quality images from undersampled acquisitions, enabling faster scans while preserving diagnostic fidelity. Recent reconstruction methods are typically trained from scratch and often require large amounts of task-specific data, limiting their robustness under data scarcity and distribution shifts. In this work, we investigate whether pretrained vision foundation models can serve as effective priors for accelerated cardiac MRI reconstruction. We propose a reconstruction framework that integrates frozen and parameter-efficiently adapted visual encoders, including CLIP, BiomedCLIP, and DINOv2, within a transformer-based reconstruction architecture. Extensive experiments on the CMRxRecon2023 and CMRxRecon2024 benchmarks demonstrate that pretrained representations consistently outperform a transformer trained from scratch across multiple acceleration factors. We further evaluate performance under limited supervision and cross-dataset transfer, showing that foundation models provide superior data efficiency and generalization. While frozen representations are particularly effective in extreme low-data regimes, Low-Rank Adaptation (LoRA) yields additional gains when moderate amounts of training data are available. Among the evaluated backbones, DINOv2 achieves the strongest overall performance. These findings highlight the potential of vision foundation models as robust and transferable priors for cardiac MRI reconstruction. Code is publicly available at: \href{https://github.com/Hashmi360/BeyondScratch-CMR}{github.com/Hashmi360/BeyondScratch-CMR}.

\end{abstract}

\section{Introduction}
\label{sec:intro}

Cardiovascular disease remains the leading cause of mortality worldwide, motivating the need for accurate, non-invasive imaging tools for early diagnosis and clinical decision-making \cite{xin2023fill, hashmi2026natural}. Cardiac magnetic resonance imaging (CMR) is widely considered a reference standard for assessing cardiac anatomy, function, and pathology, offering high-resolution soft-tissue contrast without ionizing radiation \cite{xin2023fill, eyre2022simultaneous, lyu2025state}. However, MRI acquisition is intrinsically slow because measurements are collected sequentially in k-space \cite{yiasemis2022recurrent}. Long scan times increase patient discomfort, limit scanner throughput, and make acquisitions more vulnerable to motion artifacts, which is particularly challenging in cardiac imaging \cite{yiasemis2022recurrent}. Accelerated MRI addresses this limitation by undersampling k-space to reduce acquisition time \cite{zbontar2018fastmri, yiasemis2022recurrent}, but reconstructing diagnostic-quality images from incomplete measurements is a severely ill-posed inverse problem, requiring strong priors to suppress aliasing artifacts while preserving anatomical detail \cite{fabian2022humus, lin2022vision, hashmi2026natural}.

Classical reconstruction methods incorporate handcrafted priors, such as sparsity and compressed sensing, to regularize the recovery of images from undersampled data \cite{lustig2007sparse, donoho2006compressed, heckel2024deep}. In contrast, deep learning methods learn reconstruction priors directly from data and have substantially improved both reconstruction quality and computational efficiency in accelerated MRI \cite{lin2022vision, zbontar2018fastmri, heckel2024deep, fabian2022humus}. Existing approaches operate in the image domain, k-space domain, or hybrid image-k-space formulations \cite{yiasemis2022recurrent, kim2019loraki, lin2022vision, sriram2020end, zhang2018multi, fabian2022humus, guo2021reference}. Among these, image-domain models remain attractive due to their efficiency, simplicity, and strong empirical performance \cite{lin2022vision, heckel2024deep}. CNN- and transformer-based architectures, in particular, have shown strong capacity to learn anatomical image priors and recover high-fidelity reconstructions from undersampled acquisitions \cite{fabian2022humus, lin2022vision, heckel2024deep}.



Despite these advances, deep learning-based MRI reconstruction models remain highly data-dependent and often degrade under domain shift caused by changes in scanner hardware, acquisition protocols, anatomy, or pathology \cite{li2026sd, lin2022vision}. This limits their reliability in practical clinical deployment, where large, standardized training datasets are difficult to obtain. These challenges motivate a shift from increasingly specialized reconstruction architectures toward more transferable learning paradigms \cite{li2026sd, huix2024natural}.

Given the success of transfer learning in computer vision, a central question arises for MRI reconstruction: can representations learned from large-scale visual pretraining transfer effectively to this domain, or is task-specific training still necessary? Although vision foundation models have shown strong transferability across diverse vision tasks, their role in MRI reconstruction remains underexplored. Recent studies have incorporated foundation models into reconstruction pipelines with promising results \cite{hashmi2026natural, hashmi2025can}; however, additional reconstruction modules make it difficult to isolate the contribution of the pre-trained representations themselves. Moreover, comparisons with architecturally matched transformers trained from scratch are largely missing, leaving unclear whether foundation models provide inherently stronger priors or whether similar performance can be achieved through MRI-specific training alone. This question also extends to adaptation: while parameter-efficient fine-tuning methods such as LoRA \cite{zanella2024low, zhu2024melo} are effective in downstream vision tasks, their value for MRI reconstruction, particularly under low-data and cross-dataset generalization, remains insufficiently studied.

To address these questions, we study the role of pretrained visual representations in cardiac MRI reconstruction. We incorporate three vision foundation models, CLIP \cite{radford2021learning}, BiomedCLIP \cite{zhang2023biomedclip}, and DINOv2 \cite{oquab2023dinov2}-into the encoder of a transformer-based reconstruction framework, UNETR \cite{hatamizadeh2022unetr}. This enables a systematic empirical comparison between three representation learning paradigms: (i) training transformer encoders from scratch, (ii) using frozen pretrained representations, and (iii) adapting pretrained representations via parameter-efficient fine-tuning. Through experiments spanning multiple acceleration factors, low-data regimes, and cross-dataset generalization settings, we evaluate when and how pretrained representations benefit MRI reconstruction.

Our findings show that pretrained foundation model encoders consistently outperform transformers trained from scratch, demonstrating the value of transferable visual priors for reconstruction. While parameter-efficient adaptation can yield additional improvements, its benefits depend on the target dataset and amount of available supervision. Notably, pretraining provides the largest gains in low-data settings, whereas adaptation becomes increasingly beneficial as more MRI training data become available. Together, these results offer new insights into representation transfer for MRI reconstruction and challenge the common practice of learning transformer representations entirely from scratch.






The main contributions of this work are as follows:
\begin{itemize}

\item We present, to the best of our knowledge, the first systematic study of representation transfer for cardiac MRI reconstruction, providing a systematic empirical comparison of scratch-trained transformers, frozen vision foundation models, and parameter-efficiently adapted foundation models within a unified reconstruction framework. Our findings shift the focus from learning reconstruction representations from scratch toward adapting transferable visual priors acquired through large-scale pretraining.

\item We investigate parameter-efficient adaptation for MRI reconstruction through LoRA, revealing how the relative benefits of pretraining and adaptation evolve across different supervision levels and training-data regimes.

\item We show that pretrained vision foundation models consistently yield superior reconstruction quality and stronger cross-dataset generalization than transformers trained from scratch, highlighting the value of transferable visual priors for robust accelerated cardiac MRI reconstruction.

\end{itemize}

\section{Related Work}
\label{sec:related work}
\textbf{Deep Learning for MRI Reconstruction.} Deep learning has become the dominant paradigm for accelerated MRI reconstruction, substantially outperforming traditional compressed sensing and parallel imaging approaches \cite{zbontar2018fastmri, yiasemis2022recurrent, donoho2006compressed, lustig2007sparse, larkman2007parallel, pruessmann2006encoding}. Early methods primarily relied on CNN-based architectures, particularly U-Net variants, to learn reconstruction priors directly from data \cite{wang2016accelerating, liang2020deep, ronneberger2015u, heckel2024deep}. More recently, transformer-based models have demonstrated strong performance by capturing long-range spatial dependencies and preserving fine anatomical structures \cite{lin2022vision, guo2023reconformer, guo2021reference, fabian2022humus, hatamizadeh2022unetr}. Despite these advances, transformer-based reconstruction models are often data-hungry \cite{kumar2025medvisionllama} and sensitive to distribution shifts, raising the question of whether large-scale visual pretraining can provide more transferable and data-efficient representations for MRI reconstruction.

\noindent\textbf{Foundation Models.} Vision foundation models trained on large-scale datasets have emerged as powerful sources of transferable visual representations, achieving strong performance across a wide range of downstream tasks \cite{huix2024natural}. Models such as CLIP \cite{radford2021learning}, BiomedCLIP \cite{zhang2023biomedclip}, and DINOv2 \cite{oquab2023dinov2} have demonstrated impressive transferability in both natural and medical imaging applications \cite{anand2023one, baharoon2023evaluating, muller2022joint, song2024dino}. However, their application to MRI reconstruction remains limited. Existing studies incorporate foundation models within larger reconstruction pipelines \cite{hashmi2025can, hashmi2026natural}, making it difficult to disentangle the contribution of the pretrained representations from that of the reconstruction architecture itself. Consequently, several fundamental questions remain unanswered: Do foundation models provide more effective representations than architecturally matched transformer encoders trained from scratch? How should such pretrained representations be adapted for MRI reconstruction? And under what conditions does adaptation improve reconstruction performance and generalization?


\noindent\textbf{Parameter-Efficient Adaptation of Foundation Models.} Adapting large foundation models through full fine-tuning is often computationally expensive and memory intensive \cite{lian2024less}. Parameter-efficient fine-tuning (PEFT) methods address this challenge by updating only a small subset of parameters while preserving most pretrained weights \cite{lian2024less, dutt2023parameter}. Among these, LoRA \cite{hu2022lora} has emerged as a particularly effective strategy, achieving strong performance across a variety of vision and language tasks \cite{lian2024less, dutt2023parameter, cekmeceli2024vision}. However, its role in MRI reconstruction remains poorly understood. In particular, it is unclear whether adapting pretrained visual representations consistently improves reconstruction quality, how its benefits vary with training data availability, and whether it enhances cross-dataset generalization. We address these questions in this work.

\section{Methodology}
\label{sec: method}

This work investigates the transferability and adaptation of pretrained visual representations for cardiac MRI reconstruction. To this end, we adopt a UNETR-based reconstruction framework \cite{hatamizadeh2022unetr} and perform a controlled comparison of three representation learning strategies: (i) transformer encoders trained from scratch, (ii) frozen pretrained foundation model encoders, and (iii) pretrained encoders adapted using Low-Rank Adaptation (LoRA) \cite{hu2022lora}. Specifically, we replace the original UNETR encoder with CLIP \cite{radford2021learning}, BiomedCLIP \cite{zhang2023biomedclip}, or DINOv2 \cite{oquab2023dinov2}, while keeping the decoder and reconstruction pipeline unchanged. This design isolates the effect of representation transfer from architectural factors, enabling a direct assessment of pretraining and adaptation across varying acceleration factors, training-data regimes, and cross-dataset generalization settings.

\subsection{Baseline UNETR Architecture}
\label{sec:unetr arch}

We formulate MRI reconstruction as an image-to-image translation task, where an aliased image reconstructed from undersampled k-space measurements is mapped to its fully sampled reference \cite{heckel2024deep}. As our baseline, we employ a two-dimensional UNETR architecture \cite{hatamizadeh2022unetr}, which combines a Vision Transformer (ViT) encoder \cite{vaswani2017attention, dosovitskiy2020image} with a U-Net-style decoder \cite{ronneberger2015u}. Given an undersampled MRI image $(x \in \mathbb{R}^{224 \times 224 \times 1})$, the image is partitioned into non-overlapping patches of size $(16 \times 16)$, resulting in a sequence of 196 image patches. Each patch is flattened and linearly projected into a 768-dimensional embedding space, and learnable positional embeddings are added to preserve spatial information. These tokens are processed by a six-layer ViT encoder with 12 attention heads and an MLP dimension of 3072, where each transformer block consists of multi-head self-attention and a feed-forward network with residual connections and layer normalization \cite{vaswani2017attention, dosovitskiy2020image, ba2016layer}. Intermediate representations from layers $z_3$, $z_4$, $z_5$, and $z_6$ are extracted to establish multi-scale skip connections between the encoder and decoder.

The decoder reshapes the token embeddings into spatial feature maps and progressively upsamples them through a series of transposed convolution and feature-fusion stages. The bottleneck representation ($z_6$) is successively refined using skip features from $z_5$, $z_4$, and $z_3$, enabling the recovery of both global contextual information and fine anatomical details. An additional shallow convolutional pathway extracts low-level features directly from the input image and injects them into the final decoding stage. The spatial resolution is gradually restored from $14\times14$ to $224\times224$, and a final $1\times1$ convolution produces the reconstructed MRI image. This architecture serves as the baseline framework for evaluating different representation learning strategies.

\subsection{Proposed Foundation Model Reconstruction Framework}
\label{sec:foundation model arch}

To study the impact of large-scale visual pretraining on MRI reconstruction, we replace the transformer encoder of the baseline UNETR architecture with pretrained vision foundation models while keeping the decoder, reconstruction objective, and training protocol unchanged. We evaluate three representative foundation models: CLIP (Contrastive Language-Image Pre-training)  \cite{radford2021learning}, DINOv2 (DIstillation with NO labels) \cite{oquab2023dinov2}, and BiomedCLIP \cite{zhang2023biomedclip}, which span complementary pretraining paradigms including multimodal contrastive learning, self-supervised representation learning, and biomedical vision-language pretraining. Notably, BiomedCLIP extends the CLIP framework to the biomedical domain and is pretrained on PMC-15M, a large-scale collection of biomedical image-text pairs. This controlled design isolates the contribution of pretrained representations, enabling a direct comparison between transformers trained solely on MRI data and representations acquired through large-scale visual pretraining.

Given an undersampled MRI image $x \in \mathbb{R}^{224\times224\times1}$, the input is replicated across three channels and processed by the pretrained encoder. Hidden representations from all transformer blocks are extracted, while the class token is discarded and only patch-level embeddings are retained. Rather than relying on the final transformer layer alone, we aggregate features from all encoder layers through a feature fusion module consisting of Layer Normalization followed by a two-layer MLP with GELU activation \cite{ba2016layer,cai2024vip}. The resulting fused representation serves as the bottleneck feature map for reconstruction, enabling the model to exploit both low-level structural cues and high-level semantic information learned during pretraining \cite{hashmi2025can}.

To preserve the hierarchical reconstruction strategy of UNETR, intermediate hidden states from multiple encoder depths are additionally used as skip connections and integrated into the decoder through progressive feature-fusion stages. The decoder architecture is retained without modification and progressively upsamples the fused bottleneck representation while incorporating multi-scale encoder features. An additional shallow convolutional pathway extracts low-level image information directly from the input image and injects it into the final decoding stage. While CLIP and BiomedCLIP naturally produce a $14\times14$ token grid compatible with the decoder, DINOv2 generates a $16\times16$ grid due to its patch size of 14; therefore, DINOv2 features are bilinearly interpolated to the required spatial resolution before decoding. By maintaining a common decoder architecture across all models, the proposed framework isolates the effect of representation transfer from architectural factors and enables a direct assessment of pretrained visual priors for cardiac MRI reconstruction.

\begin{figure*}[t]
    \centering
    \includegraphics[width=\textwidth]{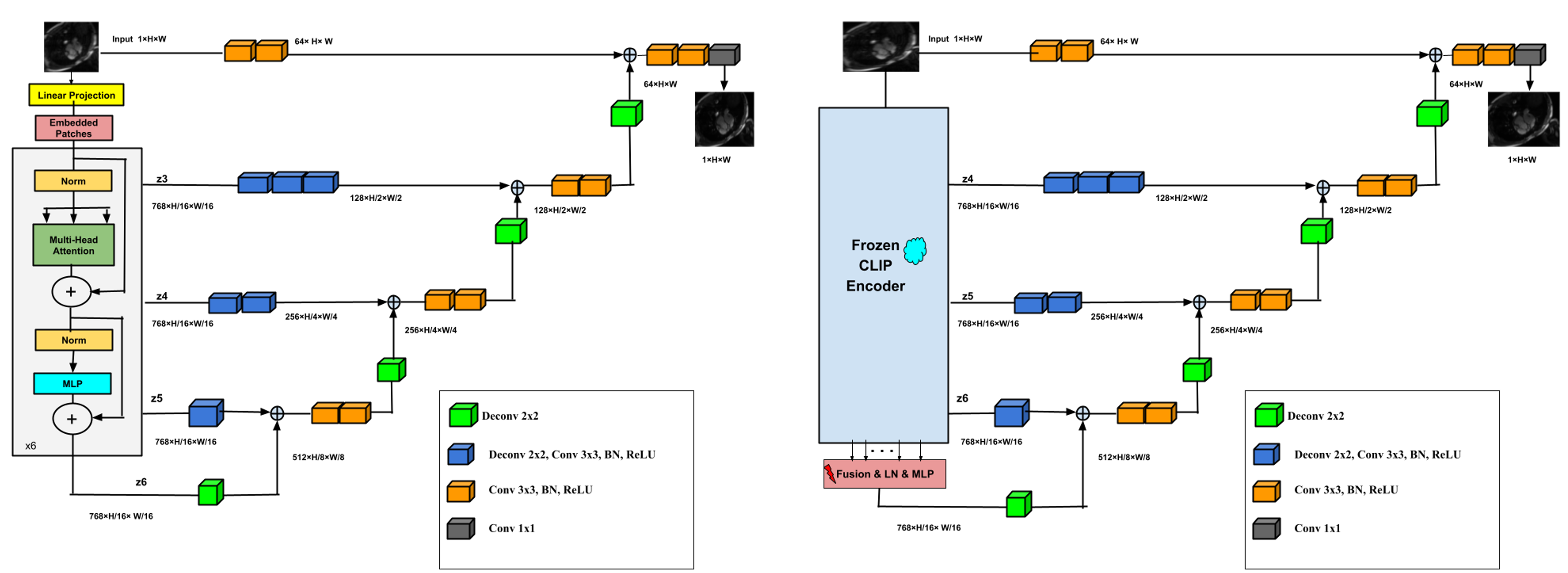}
    \caption{
    Overview of the baseline UNETR architecture (left) and the proposed foundation model reconstruction framework (right). The scratch-trained transformer encoder of UNETR is replaced with a pretrained vision foundation model (CLIP, BiomedCLIP, or DINOv2). Features from all transformer layers are fused to form the bottleneck representation, while intermediate features provide multi-scale skip connections to the decoder. The encoder is evaluated in both frozen and LoRA-adapted settings, enabling a controlled study of representation transfer for cardiac MRI reconstruction.
    }
    \label{fig:architecture}
\end{figure*}

\subsection{LoRA-Based Fine-Tuning for Efficient Adaptation}
\label{sec:lora method}

While frozen foundation model encoders provide strong transferable visual representations, their features may not be fully aligned with the characteristics of cardiac MRI data. To bridge this gap, we investigate Low-Rank Adaptation (LoRA) \cite{hu2022lora}, a parameter-efficient fine-tuning (PEFT) strategy that enables task-specific adaptation while preserving the pretrained weights. This approach is particularly attractive for medical imaging, where training data are often limited and full fine-tuning of large foundation models can be computationally expensive.

Given an input $x$, hidden state $h$, and pretrained weight matrix $\mathbf{W}\in\mathbb{R}^{d_1\times d_2}$, LoRA models the task-specific update as a low-rank decomposition,

\begin{equation}
h = \mathbf{W}x + \gamma \Delta\mathbf{W}x
= \mathbf{W}x + \gamma \mathbf{B}\mathbf{A}x,
\label{eq:lora}
\end{equation}

where $\mathbf{A}\in\mathbb{R}^{r\times d_2}$ and $\mathbf{B}\in\mathbb{R}^{d_1\times r}$ are trainable low-rank matrices, $r \ll \min(d_1,d_2)$ denotes the adaptation rank, and $\gamma$ is a scaling factor \cite{hu2022lora, zanella2024low}. Following \cite{hu2022lora}, $\mathbf{A}$ is initialized using Kaiming initialization, while $\mathbf{B}$ is initialized to zero.

In our framework, LoRA modules are inserted into the query, key, and value projection layers of every transformer block, while the original foundation model parameters remain frozen \cite{zanella2024low}. Consequently, only the LoRA parameters, token fusion module, and reconstruction decoder are optimized during training. This design enables efficient specialization of pretrained representations to MRI reconstruction and allows us to directly assess the trade-off between frozen transfer and parameter-efficient adaptation across varying data regimes and generalization settings.

\section{Experiments}
\label{sec:experiments}


We evaluate the proposed framework on multiple cardiac MRI reconstruction datasets under standard, low-data, and cross-dataset generalization settings. The experiments are designed to assess the effectiveness of representation transfer for MRI reconstruction by addressing three key questions: (i) Do pretrained foundation model representations outperform architecturally matched transformer encoders trained from scratch? (ii) Does parameter-efficient adaptation via Low-Rank Adaptation (LoRA) provide additional benefits beyond frozen pretrained representations? (iii) How do scratch-trained, frozen, and LoRA-adapted models behave under limited training data and distribution shifts across datasets?

\subsection{Datasets}
\label{sec:datasets}
\noindent\textbf{CMRxRecon2023 Dataset.}
Experiments were conducted on the CMRxRecon2023 dataset \cite{wang2023cmrxrecon}, a large-scale multi-coil cardiac MRI benchmark acquired on 3T scanners. The dataset comprises 120 fully sampled training cases covering multiple cardiac MRI contrasts, including cine, T1 mapping, and T2 mapping, together with diverse anatomical views such as short-axis and 2-, 3-, and 4-chamber acquisitions. To prevent data leakage between subjects, the cases were randomly divided at the patient level into training, validation, and test sets using a 70\%/10\%/20\% split, resulting in 2,940, 420, and 852 samples, respectively.

\noindent\textbf{CMRxRecon2024.}
To evaluate reconstruction performance across multiple cardiac MRI modalities, we additionally use the CMRxRecon2024 dataset \cite{wang2025cmrxrecon2024}, which comprises multi-coil k-space acquisitions from 200 healthy volunteers acquired on a 3T Siemens Vida scanner. The dataset includes a diverse range of cardiac MRI modalities, including cine, aorta, T1 mapping, T2 mapping, and tagging sequences, enabling the assessment of reconstruction models across varying image contrasts and anatomical structures \cite{wang2025cmrxrecon2024}. Following the same protocol as CMRxRecon2023, the data were randomly partitioned at the patient level into training, validation, and test sets using a 70\%/10\%/20\% split, resulting in 12,537, 1,803, and 3,597 samples, respectively. Experiments were performed at acceleration factors of $4\times$, $8\times$, and $10\times$.

\subsection{ Implementation Details}
\label{sec:implementation details}

All models were implemented in PyTorch and trained on an NVIDIA RTX 4090 GPU using mixed-precision training. Input MRI slices were reconstructed from undersampled multi-coil k-space data by applying an inverse Fast Fourier Transform (iFFT) to each coil, followed by Root-Sum-of-Squares (RSS) coil combination \cite{larsson2003snr,xiang2023diffcmr}. The resulting magnitude images were normalized to $[0,1]$. For foundation-model-based approaches, images were center-cropped or padded to $224\times224$ and replicated across three channels to match the input format of the pretrained encoders.

The scratch-trained UNETR baseline used a 2D ViT encoder with patch size $16$, hidden dimension $768$, six transformer layers, 12 attention heads and MLP dimension $3072$. It was optimized using AdamW with a learning rate of $1\times10^{-4}$ and weight decay of $1\times10^{-2}$. For frozen foundation model experiments, CLIP, BiomedCLIP, and DINOv2 (ViT-B) encoders were initialized from publicly available checkpoints and kept fixed, while only the feature fusion layers and reconstruction decoder were trained using AdamW with a learning rate of $2\times10^{-4}$.

For parameter-efficient adaptation, LoRA modules were inserted into the transformer attention projection layers, while the original pretrained encoder weights remained frozen. Unless otherwise stated, LoRA-based models were trained using AdamW with a learning rate of $1\times10^{-4}$ and weight decay of $1\times10^{-2}$. We further performed LoRA rank ablations with $r\in\{8,64,128\}$ across acceleration factors to assess the effect of adaptation capacity. During LoRA training, only the LoRA parameters, feature fusion layers, and reconstruction decoder were updated.

All models were trained for up to 300 epochs with a batch size of 8. The learning rate was reduced by a factor of $0.5$ if the validation loss did not improve for 5 epochs, and early stopping was applied after 15 epochs without improvement. The Structural Similarity Index Measure (SSIM) \cite{wang2003multiscale} loss was used as the training objective \cite{sriram2020end, xin2023fill}. Reconstruction quality was evaluated using PSNR, SSIM, and NMSE. All reported results are computed on the held-out test set using the checkpoint with the best validation performance.

\section{Results}
\subsection{Comparison with Scratch-Trained Transformers}
\label{sec:frozen_results}

We first investigate whether representations acquired through large-scale visual pretraining provide advantages over transformer encoders trained solely on MRI data. Table~\ref{tab:main_results_1} compares the scratch-trained UNETR baseline with frozen CLIP, BiomedCLIP, and DINOv2 encoders on CMRxRecon2023 and CMRxRecon2024 across acceleration factors of $4\times$, $8\times$, and $10\times$.

Across all datasets and acceleration factors, frozen foundation model encoders consistently outperform the scratch-trained UNETR baseline. Remarkably, these improvements are achieved with approximately 49\% fewer trainable parameters (33.7M vs. 65.7M), highlighting the effectiveness of large-scale visual pretraining as a transferable reconstruction prior. The gains become increasingly evident at higher acceleration factors, where stronger priors are required to recover anatomical structures from severely undersampled measurements.

Among the evaluated foundation models, DINOv2 achieves the strongest overall performance. On CMRxRecon2023, Frozen DINOv2 obtains the best results at $4\times$ and $8\times$ acceleration, achieving SSIM values of 0.9073 and 0.8787 respectively. At $10\times$ acceleration, Frozen CLIP marginally outperforms DINOv2, achieving the highest SSIM (0.8675), PSNR (30.06 dB), and lowest NMSE (0.0538). Similar trends are observed on CMRxRecon2024, where DINOv2 achieves the highest SSIM and PSNR at both $4\times$ and $8\times$ acceleration, while CLIP attains the lowest NMSE at $4\times$ and the strongest PSNR and NMSE at $10\times$.

These results suggest that pretrained visual representations capture structural and semantic information that remains highly relevant for MRI reconstruction. In particular, the strong performance of DINOv2 indicates that self-supervised visual pretraining produces representations that generalize effectively to inverse imaging tasks. Overall, the findings show that large-scale visual pretraining provides a substantially stronger initialization than learning transformer representations exclusively from the available MRI training data, supporting the use of foundation models as transferable reconstruction priors.

\begin{table*}[t]
\centering
\scriptsize
\setlength{\tabcolsep}{3pt}
\renewcommand{\arraystretch}{1.10}
\caption{
Quantitative reconstruction performance on CMRxRecon2023 and CMRxRecon2024. Best results are shown in bold.
}
\label{tab:main_results_1}

\begin{tabular}{p{0.6cm}p{3.6cm}|ccc|ccc}
\toprule
\multirow{2}{*}{Acc.} &
\multirow{2}{*}{Model} 
& \multicolumn{3}{c|}{CMRxRecon2023}
& \multicolumn{3}{c}{CMRxRecon2024} \\
\cmidrule(lr){3-5}
\cmidrule(lr){6-8}
&
& SSIM$\uparrow$ & PSNR$\uparrow$ & NMSE$\downarrow$
& SSIM$\uparrow$ & PSNR$\uparrow$ & NMSE$\downarrow$ \\
\midrule

\multirow{4}{*}{4$\times$}
& UNETR [65.7M]
& 0.8904 & 31.1405 & 0.0458
& 0.8801 & 30.7491 & 0.0552 \\

& Frozen CLIP [33.7M]
& 0.9035 & 32.1443 & \textbf{0.0344}
& 0.8950 & 31.4805 & \textbf{0.0448} \\

& Frozen BiomedCLIP [33.7M]
& 0.9027 & 32.1010 & 0.0351
& 0.8911 & 31.1441 & 0.0495 \\

& Frozen DINOv2 [33.7M]
& \textbf{0.9073} & \textbf{32.2097} & 0.0348
& \textbf{0.8979} & \textbf{31.5111} & 0.0456 \\

\midrule

\multirow{4}{*}{8$\times$}
& UNETR
& 0.8625 & 29.7299 & 0.0632
& 0.8412 & 29.0134 & 0.0831 \\

& Frozen CLIP
& 0.8770 & 30.6202 & 0.0490
& 0.8553 & 29.5663 & 0.0691 \\

& Frozen BiomedCLIP
& 0.8738 & 30.3092 & 0.0552
& 0.8505 & 29.5788 & 0.0685 \\

& Frozen DINOv2
& \textbf{0.8787} & \textbf{30.7081} & \textbf{0.0480}
& \textbf{0.8606} & \textbf{29.9150} & \textbf{0.0638} \\

\midrule

\multirow{4}{*}{10$\times$}
& UNETR
& 0.8524 & 29.1022 & 0.0765
& 0.8343 & 28.5577 & 0.0890 \\

& Frozen CLIP
& \textbf{0.8675} & \textbf{30.0675} & \textbf{0.0538}
& 0.8468 & \textbf{29.2754} & \textbf{0.0730} \\

& Frozen BiomedCLIP
& 0.8636 & 29.5681 & 0.0640
& 0.8419 & 28.8263 & 0.0846 \\

& Frozen DINOv2
& 0.8672 & 29.9788 & 0.0573
& \textbf{0.8514} & 29.1639 & 0.0768 \\

\bottomrule
\end{tabular}
\end{table*}

\newcommand{\up}{\textcolor{red}{\uparrow}}
\newcommand{\down}{\textcolor{green!50!black}{\downarrow}}

\begin{table*}[t]
\centering
\caption{Effect of LoRA adaptation on pretrained foundation models across CMRxRecon2023 and CMRxRecon2024. Arrows indicate the change relative to the corresponding frozen model: green arrows ($\downarrow$) denote performance degradation, while red arrows ($\uparrow$) denote an improvement. Best results for each acceleration factor and dataset are shown in bold.}
\label{tab:lora_results}
\scriptsize
\setlength{\tabcolsep}{3pt}
\renewcommand{\arraystretch}{1.10}
\begin{tabular}{p{0.6cm}p{2.5cm}|ccc|ccc}
\toprule
\multirow{2}{*}{Acc.} &
\multirow{2}{*}{Model}
& \multicolumn{3}{c|}{CMRxRecon2023}
& \multicolumn{3}{c}{CMRxRecon2024} \\
\cmidrule(lr){3-5}
\cmidrule(lr){6-8}
&
& SSIM$\uparrow$ & PSNR$\uparrow$ & NMSE$\downarrow$
& SSIM$\uparrow$ & PSNR$\uparrow$ & NMSE$\downarrow$ \\
\midrule

\multirow{6}{*}{4$\times$}
& Frozen CLIP      & 0.9035 & 32.1443 & 0.0344 & 0.8950 & 31.4805 & 0.0448 \\
& LoRA CLIP        & 0.9013{\tiny$^{\down}$} & 32.1458 {\tiny$^{\up}$} & 0.0340{\tiny$^{\up}$} & 0.9003{\tiny$^{\up}$} & 32.2644{\tiny$^{\up}$} & 0.0377{\tiny$^{\up}$} \\

& Frozen BiomedCLIP & 0.9027 & 32.1010 & 0.0351 & 0.8911 & 31.1441 & 0.0495 \\
& LoRA BiomedCLIP   & 0.8972{\tiny$^{\down}$} & 31.9133{\tiny$^{\down}$} & 0.0352{\tiny$^{\down}$} & 0.8945{\tiny$^{\up}$} & 31.8295{\tiny$^{\up}$} & 0.0407{\tiny$^{\up}$} \\

& Frozen DINOv2    & \textbf{0.9073} & 32.2097 & 0.0348 & 0.8979 & 31.5111 & 0.0456 \\
& LoRA DINOv2      & 0.9061{\tiny$^{\down}$} & \textbf{32.4175}{\tiny$^{\up}$} & \textbf{0.0319}{\tiny$^{\up}$} & \textbf{0.9054}{\tiny$^{\up}$} & \textbf{32.5112}{\tiny$^{\up}$} & \textbf{0.0353}{\tiny$^{\up}$} \\

\midrule

\multirow{6}{*}{8$\times$}
& Frozen CLIP      & 0.8770 & 30.6202 & 0.0490 & 0.8553 & 29.5663 & 0.0691 \\
& LoRA CLIP        & 0.8710{\tiny$^{\down}$} & 30.4831{\tiny$^{\down}$} & 0.0492{\tiny$^{\down}$} & 0.8591{\tiny$^{\up}$} & 30.2022{\tiny$^{\up}$} & 0.0585{\tiny$^{\up}$} \\

& Frozen BiomedCLIP & 0.8738 & 30.3092 & 0.0552 & 0.8505 & 29.5788 & 0.0685 \\
& LoRA BiomedCLIP   & 0.8706{\tiny$^{\down}$} & 30.4793{\tiny$^{\up}$} & 0.0516{\tiny$^{\up}$} & 0.8543{\tiny$^{\up}$} & 29.9158{\tiny$^{\up}$} & 0.0636{\tiny$^{\up}$} \\

& Frozen DINOv2    & \textbf{0.8787} & 30.7081 & 0.0480 & 0.8606 & 29.9150 & 0.0638 \\
& LoRA DINOv2      & 0.8776{\tiny$^{\down}$} & \textbf{30.9313}{\tiny$^{\up}$} & \textbf{0.0441}{\tiny$^{\up}$} & \textbf{0.8663}{\tiny$^{\up}$} & \textbf{30.5859}{\tiny$^{\up}$} & \textbf{0.0543}{\tiny$^{\up}$} \\

\midrule

\multirow{6}{*}{10$\times$}
& Frozen CLIP      & \textbf{0.8675} & 30.0675 & 0.0538 & 0.8468 & 29.2754 & 0.0730 \\
& LoRA CLIP        & 0.8629{\tiny$^{\down}$} & 29.9360{\tiny$^{\down}$} & 0.0549{\tiny$^{\down}$} & 0.8500{\tiny$^{\up}$} & 29.5561{\tiny$^{\up}$} & 0.0675{\tiny$^{\up}$} \\

& Frozen BiomedCLIP & 0.8636 & 29.5681 & 0.0640 & 0.8419 & 28.8263 & 0.0846 \\
& LoRA BiomedCLIP   & 0.8605{\tiny$^{\down}$} & 29.6505{\tiny$^{\up}$} & 0.0605{\tiny$^{\up}$} & 0.8459{\tiny$^{\up}$} & 29.3922{\tiny$^{\up}$} & 0.0709{\tiny$^{\up}$} \\

& Frozen DINOv2    & 0.8672 & 29.9788 & 0.0573 & 0.8514 & 29.1639 & 0.0768 \\
& LoRA DINOv2      & 0.8649{\tiny$^{\down}$} & \textbf{30.1798}{\tiny$^{\up}$} & \textbf{0.0511}{\tiny$^{\up}$} & \textbf{0.8571}{\tiny$^{\up}$} & \textbf{29.8925}{\tiny$^{\up}$} & \textbf{0.0635}{\tiny$^{\up}$} \\

\bottomrule
\end{tabular}
\end{table*}

\subsection{Adapting Pretrained Representations with LoRA}
\label{sec:lora_results}

To assess whether pretrained representations can be further specialized for MRI reconstruction, we apply LoRA to the frozen foundation model encoders. A rank sensitivity study was performed using $r\in\{8,64,128\}$ across all foundation models, datasets, and acceleration factors. The optimal rank was selected independently for each configuration based on the overall trade-off between SSIM, PSNR, and NMSE. The selected ranks varied across models and reconstruction settings. CLIP favored $r=128$ at $4\times$ and $8\times$ acceleration and $r=64$ at $10\times$ on CMRxRecon2023, while on CMRxRecon2024 it selected $r=128$ at $4\times$ and $r=64$ at $8\times$ and $10\times$. BiomedCLIP consistently selected $r=128$ across all acceleration factors and datasets. DINOv2 favored $r=128$ at $4\times$ and $8\times$ on both datasets, as well as at $10\times$ on CMRxRecon2024, with $r=64$ selected only at $10\times$ on CMRxRecon2023. The results reported in Table~\ref{tab:lora_results} correspond to the best-performing rank for each configuration, while the complete rank ablation study and associated trainable parameter counts are provided in the supplementary material.

Table~\ref{tab:lora_results} compares frozen and LoRA-adapted variants of CLIP, BiomedCLIP, and DINOv2. Overall, the effectiveness of LoRA is strongly dataset dependent. On CMRxRecon2024, adaptation consistently improves reconstruction quality across all foundation models, yielding higher PSNR and SSIM together with lower NMSE. The largest gains are observed for DINOv2, where LoRA improves performance at every acceleration factor and achieves the best overall results. 

In contrast, the benefits of LoRA on CMRxRecon2023 are more modest. While adaptation consistently improves PSNR and NMSE, it often results in small reductions in SSIM relative to the frozen counterparts, suggesting that the pretrained representations are already well aligned with the reconstruction task. Notably, DINOv2 benefits most consistently from adaptation, achieving improved PSNR and NMSE across all acceleration factors and datasets. 
These findings indicate that while frozen pretraining provides strong reconstruction priors, parameter-efficient adaptation can further specialize pretrained representations when sufficient data diversity and supervision are available. The larger gains on CMRxRecon2024 suggest that LoRA is most effective when adapting to richer and larger-scale training datasets.

\begin{table*}[t]
\centering
\scriptsize
\setlength{\tabcolsep}{3pt}
\renewcommand{\arraystretch}{1.10}
\caption{
Low-data regime analysis at $10\times$ acceleration on CMRxRecon2023 and CMRxRecon2024. Results are reported using 5\%, 25\%, 50\%, and 75\% of the available training data. Best results within each data regime are highlighted in bold.
}
\label{tab:low_data}
\begin{tabular}{c|l|ccc|ccc}
\toprule
\multirow{2}{*}{Data}
&
\multirow{2}{*}{Model}
&
\multicolumn{3}{c|}{CMRxRecon2023}
&
\multicolumn{3}{c}{CMRxRecon2024}
\\
\cmidrule(lr){3-5}
\cmidrule(lr){6-8}
&
&
SSIM$\uparrow$
&
PSNR$\uparrow$
&
NMSE$\downarrow$
&
SSIM$\uparrow$
&
PSNR$\uparrow$
&
NMSE$\downarrow$
\\
\midrule

\multirow{7}{*}{5\%}
& UNETR             & 0.7900 & 27.1943 & 0.1098 & 0.7803 & 26.6047 & 0.1445 \\
& Frozen CLIP       & 0.8310 & 28.2532 & 0.0835 & 0.8116 & 27.8297 & 0.1007 \\
& LoRA CLIP         & 0.8196 & 27.7251 & 0.0977 & 0.8061 & 27.7494 & 0.0991 \\
& Frozen BiomedCLIP & 0.8199 & 27.7550 & 0.0970 & 0.8056 & 27.5921 & 0.1055 \\
& LoRA BiomedCLIP   & 0.8168 & 27.5929 & 0.1025 & 0.8011 & 26.8955 & 0.1346 \\
& Frozen DINOv2     & \textbf{0.8348} & \textbf{28.4418} & \textbf{0.0825} & \textbf{0.8175} & \textbf{27.9996} & \textbf{0.0942} \\
& LoRA DINOv2       & 0.8242 & 27.6067 & 0.1037 & 0.8094 & 27.6240 & 0.0995 \\
\midrule

\multirow{7}{*}{25\%}
& UNETR             & 0.8335 & 28.4024 & 0.0822 & 0.8168 & 27.9679 & 0.1031 \\
& Frozen CLIP       & 0.8562 & \textbf{29.5982} & \textbf{0.0609} & 0.8326 & 28.6022 & 0.0860 \\
& LoRA CLIP         & 0.8480 & 29.2490 & 0.0644 & 0.8339 & 28.8068 & 0.0801 \\
& Frozen BiomedCLIP & 0.8523 & 29.2954 & 0.0676 & 0.8298 & 28.5149 & 0.0874 \\
& LoRA BiomedCLIP   & 0.8484 & 29.1103 & 0.0700 & 0.8284 & 28.6318 & 0.0836 \\
& Frozen DINOv2     & \textbf{0.8591} & 29.1801 & 0.0709 & 0.8375 & 28.6793 & 0.0846 \\
& LoRA DINOv2       & 0.8521 & 29.1974 & 0.0639 & \textbf{0.8380} & \textbf{29.0067} & \textbf{0.0765} \\
\midrule

\multirow{7}{*}{50\%}
& UNETR             & 0.8442 & 28.7257 & 0.0791 & 0.8268 & 28.3097 & 0.0941 \\
& Frozen CLIP       & 0.8630 & 29.8422 & 0.0573 & 0.8399 & 28.8425 & 0.0814 \\
& LoRA CLIP         & 0.8578 & 29.5904 & 0.0601 & 0.8428 & 29.3343 & 0.0704 \\
& Frozen BiomedCLIP & 0.8596 & 29.5823 & 0.0644 & 0.8372 & 28.8198 & 0.0806 \\
& LoRA BiomedCLIP   & 0.8549 & 29.6632 & 0.0585 & 0.8366 & 29.0000 & 0.0762 \\
& Frozen DINOv2     & \textbf{0.8661} & \textbf{30.0347} & \textbf{0.0555} & 0.8461 & 29.1757 & 0.0731 \\
& LoRA DINOv2       & 0.8581 & 29.2194 & 0.0678 & \textbf{0.8493} & \textbf{29.5765} & \textbf{0.0677} \\
\midrule

\multirow{7}{*}{75\%}
& UNETR             & 0.8503 & 29.1969 & 0.0686 & 0.8286 & 28.3560 & 0.0955 \\
& Frozen CLIP       & 0.8663 & 29.8987 & 0.0577 & 0.8454 & 29.1511 & 0.0757 \\
& LoRA CLIP         & 0.8620 & 29.9978 & 0.0541 & 0.8480 & 29.4933 & 0.0695 \\
& Frozen BiomedCLIP & 0.8626 & 29.8418 & 0.0580 & 0.8403 & 29.0155 & 0.0784 \\
& LoRA BiomedCLIP   & 0.8587 & 29.7396 & 0.0595 & 0.8433 & 29.1320 & 0.0772 \\
& Frozen DINOv2     & \textbf{0.8690} & \textbf{30.1899} & 0.0531 & 0.8503 & 29.3179 & 0.0726 \\
& LoRA DINOv2       & 0.8647 & 30.1706 & \textbf{0.0517} & \textbf{0.8550} & \textbf{29.5808} & \textbf{0.0684} \\
\bottomrule
\end{tabular}
\end{table*}

\subsection{Data Efficiency Under Limited Supervision}
\label{sec:low_data}

Table~\ref{tab:low_data} reports reconstruction performance at $10\times$ acceleration using only 5\%, 25\%, 50\%, and 75\% of the available training data. Across both CMRxRecon2023 and CMRxRecon2024, pretrained foundation models consistently outperform the scratch-trained UNETR baseline, demonstrating substantially greater data efficiency. The largest gains occur in the extreme 5\% regime, where Frozen DINOv2 achieves SSIM values of 0.8348 and 0.8175 on CMRxRecon2023 and CMRxRecon2024, compared with 0.7900 and 0.7803 for UNETR. These results indicate that large-scale visual pretraining provides strong reconstruction priors that remain effective under severe supervision constraints.

The effectiveness of LoRA adaptation (configured with the best-performing LoRA rank) depends strongly on the amount of available training data. At 5\% supervision, adapting the pretrained encoders generally reduces performance relative to their frozen counterparts, suggesting that the limited data are insufficient to reliably specialize the pretrained representations. In contrast, as more training data become available, the benefits of adaptation become increasingly apparent. This trend is most evident on CMRxRecon2024, where LoRA-adapted DINOv2 achieves the best overall performance at 25\%, 50\%, and 75\% training data, reaching an SSIM of 0.8550 and a PSNR of 29.58\,dB at 75\%.

Overall, the results reveal a clear transition between transfer and adaptation. Frozen foundation models provide the strongest inductive bias in highly data-constrained settings, whereas parameter-efficient adaptation becomes progressively more beneficial as supervision increases. Among all evaluated backbones, DINOv2 consistently delivers the strongest performance, highlighting the effectiveness of self-supervised visual representations for cardiac MRI reconstruction.

\begin{figure*}[t]
    \centering
    \includegraphics[width=\textwidth]{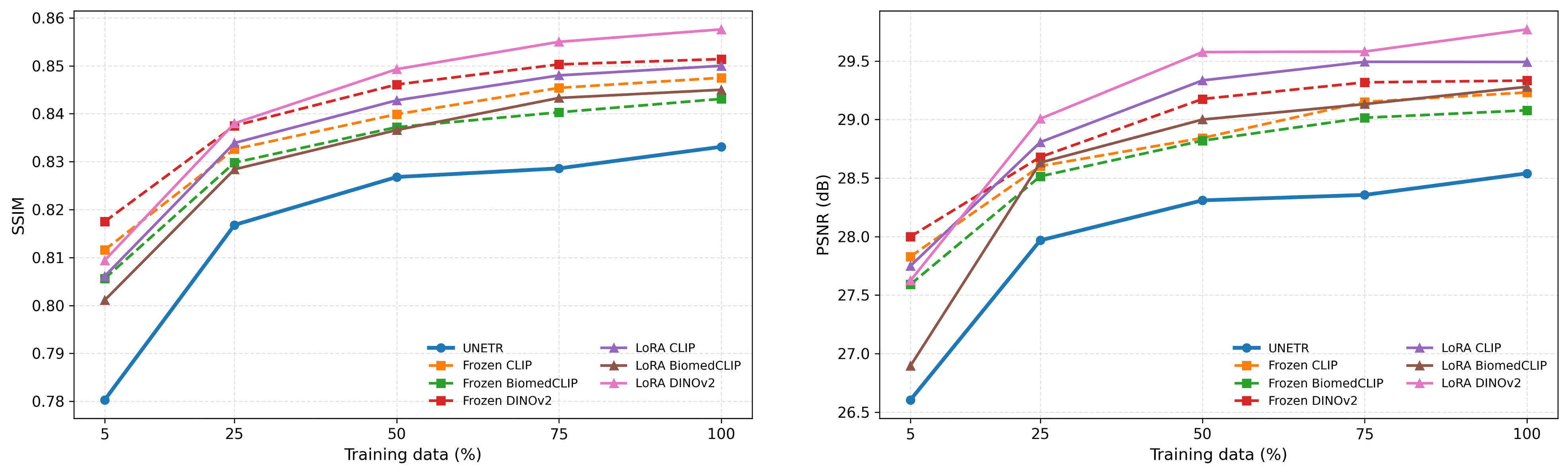}
    \caption{
    Low-data regime analysis on CMRxRecon2024 at $10\times$ acceleration. Foundation model-based methods consistently outperform UNETR across all training fractions. Frozen DINOv2 achieves the strongest performance in the extreme 5\% regime, while LoRA-adapted DINOv2 becomes increasingly advantageous as more training data becomes available.
    }
    \label{fig:architecture}
\end{figure*}

\subsection{Cross-Dataset Generalization}
\label{sec:generalization}

To evaluate robustness under distribution shift, we train on one challenge dataset and directly evaluate on the other without any additional fine-tuning. Specifically, models trained on CMRxRecon2024 are tested on CMRxRecon2023, and vice versa. Results are summarized in Table~\ref{tab:cross_dataset}.

Across both transfer directions, foundation model-based approaches consistently outperform the scratch-trained UNETR baseline, indicating that large-scale visual pretraining yields more transferable reconstruction representations. The strongest performance is achieved by DINOv2, with LoRA-adapted DINOv2 obtaining the best overall results when transferring from CMRxRecon2024 to CMRxRecon2023 (SSIM 0.8634, PSNR 29.60\,dB, NMSE 0.0583). Similar trends are observed in the reverse direction, where DINOv2 again achieves the highest reconstruction quality across the evaluated metrics.

The transfer from CMRxRecon2023 to CMRxRecon2024 proves substantially more challenging than the reverse setting, resulting in a performance drop for all methods. We attribute this asymmetry to the greater diversity and scale of CMRxRecon2024, which presents a more demanding target distribution. Nevertheless, foundation model encoders maintain a clear advantage over UNETR, demonstrating improved robustness to domain shifts.

Unlike the extreme low-data regime, where frozen representations often performed best, LoRA adaptation generally improves cross-dataset transferability. This suggests that parameter-efficient adaptation can better align pretrained representations with the reconstruction task while preserving the broad visual priors acquired during pretraining.

Overall, these results reinforce the central finding of this work: pretrained foundation models learn substantially more transferable representations for cardiac MRI reconstruction than transformers trained from scratch. Among the evaluated backbones, DINOv2 consistently exhibits the strongest generalization capability, while LoRA further enhances robustness under distribution shift.

\begin{table*}[t]
\centering
\scriptsize
\setlength{\tabcolsep}{3pt}
\renewcommand{\arraystretch}{1.10}
\caption{
Cross-dataset generalization results at $10\times$ acceleration. Models are trained on one challenge dataset and evaluated on the other dataset. Best results for each transfer direction are shown in bold.
}
\label{tab:cross_dataset}
\begin{tabular}{l|ccc|ccc}
\toprule
\multirow{2}{*}{Model}
&
\multicolumn{3}{c|}{CMRxRecon2024 $\rightarrow$ CMRxRecon2023}
&
\multicolumn{3}{c}{CMRxRecon2023 $\rightarrow$ CMRxRecon2024}
\\
\cmidrule(lr){2-4}
\cmidrule(lr){5-7}
&
SSIM$\uparrow$
&
PSNR$\uparrow$
&
NMSE$\downarrow$
&
SSIM$\uparrow$
&
PSNR$\uparrow$
&
NMSE$\downarrow$
\\
\midrule

UNETR
& 0.8293 & 27.9917 & 0.0849
& 0.7596 & 25.8888 & 0.1833 \\

Frozen CLIP
& 0.8510 & 28.8955 & 0.0686
& 0.7971 & 27.2573 & 0.1183 \\

LoRA CLIP
& 0.8533 & 29.1395 & 0.0639
& 0.7919 & 27.3316 & 0.1117 \\

Frozen BiomedCLIP
& 0.8360 & 28.2405 & 0.0796
& 0.7780 & 26.1926 & 0.1569 \\

LoRA BiomedCLIP
& 0.8473 & 28.3863 & 0.0822
& 0.7889 & 27.2543 & 0.1155 \\

Frozen DINOv2
& 0.8603 & 28.8937 & 0.0703
& \textbf{0.8016} & 27.2070 & 0.1253 \\

LoRA DINOv2
& \textbf{0.8634} & \textbf{29.6007} & \textbf{0.0583}
& 0.7953 & \textbf{27.6952} & \textbf{0.1012} \\

\bottomrule
\end{tabular}
\end{table*}

\subsection{Qualitative Evaluation}
\label{sec:qualitative}



\begin{figure*}[t]
    \centering
    \includegraphics[width=\textwidth]{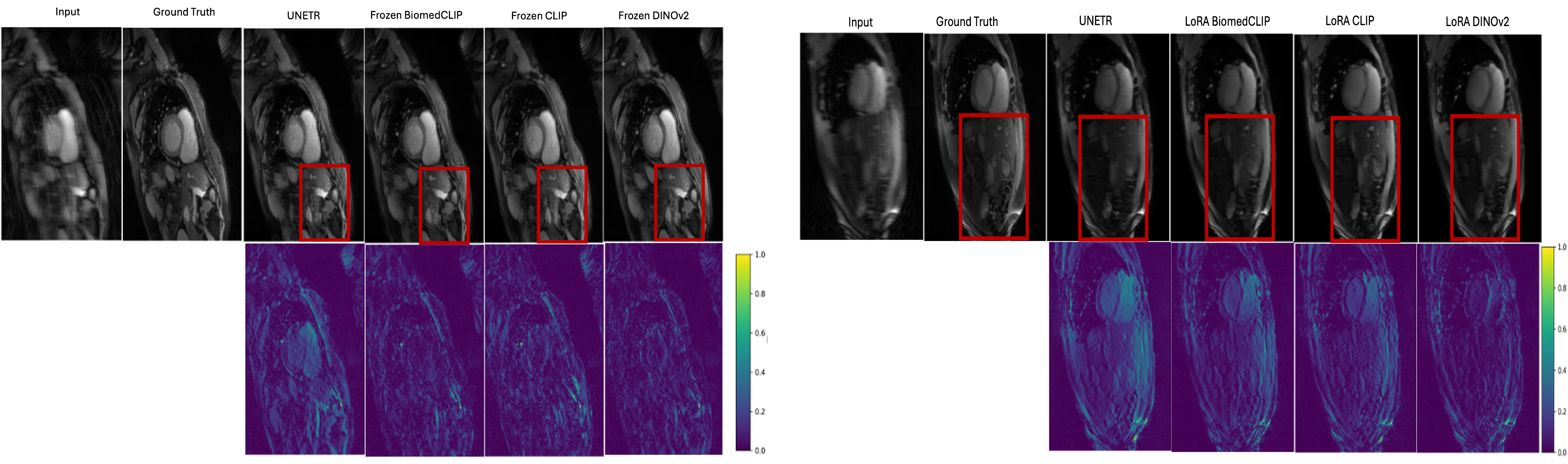}
    \caption{
    Qualitative reconstruction results and absolute error maps. Left: In-domain performance on CMRxRecon2024 at $4\times$ acceleration, where foundation models produce sharper reconstructions and lower errors than UNETR. Right: Cross-dataset transfer from CMRxRecon2024 to CMRxRecon2023 at $10\times$ acceleration, showing improved robustness and anatomical fidelity for LoRA-adapted foundation models. Red boxes indicate representative regions of interest.}
    \label{fig:qualitative}
\end{figure*}

Representative reconstructions are shown in Fig.~\ref{fig:qualitative}. Consistent with the quantitative results, foundation model-based approaches produce reconstructions with improved anatomical fidelity and fewer residual artifacts than UNETR. The differences are most evident around fine cardiac structures where pretrained encoders preserve sharper details and yield lower reconstruction errors.

Among the evaluated methods, DINOv2-based models most closely match the ground truth, exhibiting reduced error maps and improved structural consistency. The advantages of pretrained representations become particularly apparent under challenging conditions, including high acceleration factors and cross-dataset evaluation, where UNETR tends to exhibit oversmoothing and loss of fine anatomical detail. These qualitative results further support the effectiveness and transferability of foundation model representations for cardiac MRI reconstruction.

\section{Conclusion}


We presented a systematic study of representation transfer for accelerated cardiac MRI reconstruction, comparing scratch-trained transformers, frozen foundation models, and LoRA-adapted foundation models within a unified framework. Across the CMRxRecon2023 and CMRxRecon2024 benchmarks, pretrained foundation models consistently achieved superior reconstruction quality, greater data efficiency, and stronger cross-dataset generalization than a transformer trained from scratch. Our results further revealed that frozen representations are most effective in highly data-constrained settings, while parameter-efficient adaptation becomes increasingly beneficial as more task-specific supervision becomes available. Among the evaluated backbones, DINOv2 consistently delivered the strongest performance, highlighting the effectiveness of self-supervised visual representations for reconstruction. Overall, these findings demonstrate that large-scale visual pretraining provides powerful and transferable priors for cardiac MRI reconstruction and represents a promising direction for developing more data-efficient and generalizable medical image reconstruction systems.

\paragraph{Limitations.}
Our evaluation relies on average SSIM and PSNR without radiological assessment;
therefore, the observed improvements reflect empirical trends and should not be
interpreted as evidence of clinical superiority.

\section*{Acknowledgements}
This work was supported by Taighde \'Eireann–Research Ireland through the Research Ireland Centre for Research Training in Machine Learning under Grant numbers 18/CRT/6183 \& 12/RC/2289 P2.



%
%
\bibliographystyle{splncs04}
\bibliography{main}

@String(ICLR  = {Int. Conf. Learn. Represent.})

@String(ICLR  = {ICLR})

@inproceedings{xin2023fill,
  title={Fill the k-space and refine the image: Prompting for dynamic and multi-contrast MRI reconstruction},
  author={Xin, Bingyu and Ye, Meng and Axel, Leon and Metaxas, Dimitris N},
  booktitle={International Workshop on Statistical Atlases and Computational Models of the Heart},
  pages={261--273},
  year={2023},
  organization={Springer}
}

@inproceedings{hashmi2026natural,
  title={Are Natural-Domain Foundation Models Effective for Accelerated Cardiac MRI Reconstruction?},
  author={Hashmi, Anam and Maniparambil, Mayug and Dietlmeier, Julia and Curran, Kathleen M and E O'Connor, Noel},
  booktitle={Proceedings of the IEEE/CVF Conference on Computer Vision and Pattern Recognition},
  pages={6202--6211},
  year={2026}
}

@article{eyre2022simultaneous,
  title={Simultaneous multi-parametric acquisition and reconstruction techniques in cardiac magnetic resonance imaging: basic concepts and status of clinical development},
  author={Eyre, Katerina and Lindsay, Katherine and Razzaq, Saad and Chetrit, Michael and Friedrich, Matthias},
  journal={Frontiers in cardiovascular medicine},
  volume={9},
  pages={953823},
  year={2022},
  publisher={Frontiers Media SA}
}

@article{lyu2025state,
  title={The state-of-the-art in cardiac mri reconstruction: Results of the cmrxrecon challenge in miccai 2023},
  author={Lyu, Jun and Qin, Chen and Wang, Shuo and Wang, Fanwen and Li, Yan and Wang, Zi and Guo, Kunyuan and Ouyang, Cheng and T{\"a}nzer, Michael and Liu, Meng and others},
  journal={Medical Image Analysis},
  volume={101},
  pages={103485},
  year={2025},
  publisher={Elsevier}
}

@inproceedings{yiasemis2022recurrent,
  title={Recurrent variational network: a deep learning inverse problem solver applied to the task of accelerated MRI reconstruction},
  author={Yiasemis, George and Sonke, Jan-Jakob and S{\'a}nchez, Clarisa and Teuwen, Jonas},
  booktitle={Proceedings of the IEEE/CVF conference on computer vision and pattern recognition},
  pages={732--741},
  year={2022}
}

@article{zbontar2018fastmri,
  title={fastMRI: An open dataset and benchmarks for accelerated MRI},
  author={Zbontar, Jure and Knoll, Florian and Sriram, Anuroop and Murrell, Tullie and Huang, Zhengnan and Muckley, Matthew J and Defazio, Aaron and Stern, Ruben and Johnson, Patricia and Bruno, Mary and others},
  journal={arXiv preprint arXiv:1811.08839},
  year={2018}
}

@article{fabian2022humus,
  title={Humus-net: Hybrid unrolled multi-scale network architecture for accelerated mri reconstruction},
  author={Fabian, Zalan and Tinaz, Berk and Soltanolkotabi, Mahdi},
  journal={Advances in Neural Information Processing Systems},
  volume={35},
  pages={25306--25319},
  year={2022}
}

@inproceedings{lin2022vision,
  title={Vision transformers enable fast and robust accelerated MRI},
  author={Lin, Kang and Heckel, Reinhard},
  booktitle={International Conference on medical imaging with deep learning},
  pages={774--795},
  year={2022},
  organization={PMLR}
}

@article{lustig2007sparse,
  title={Sparse MRI: The application of compressed sensing for rapid MR imaging},
  author={Lustig, Michael and Donoho, David and Pauly, John M},
  journal={Magnetic Resonance in Medicine: An Official Journal of the International Society for Magnetic Resonance in Medicine},
  volume={58},
  number={6},
  pages={1182--1195},
  year={2007},
  publisher={Wiley Online Library}
}

@article{heckel2024deep,
  title={Deep learning for accelerated and robust MRI reconstruction},
  author={Heckel, Reinhard and Jacob, Mathews and Chaudhari, Akshay and Perlman, Or and Shimron, Efrat},
  journal={Magnetic Resonance Materials in Physics, Biology and Medicine},
  volume={37},
  number={3},
  pages={335--368},
  year={2024},
  publisher={Springer}
}

@article{donoho2006compressed,
  title={Compressed sensing},
  author={Donoho, David L},
  journal={IEEE Transactions on information theory},
  volume={52},
  number={4},
  pages={1289--1306},
  year={2006},
  publisher={IEEE}
}

@article{kim2019loraki,
  title={LORAKI: Autocalibrated recurrent neural networks for autoregressive MRI reconstruction in k-space},
  author={Kim, Tae Hyung and Garg, Pratyush and Haldar, Justin P},
  journal={arXiv preprint arXiv:1904.09390},
  year={2019}
}

@inproceedings{sriram2020end,
  title={End-to-end variational networks for accelerated MRI reconstruction},
  author={Sriram, Anuroop and Zbontar, Jure and Murrell, Tullie and Defazio, Aaron and Zitnick, C Lawrence and Yakubova, Nafissa and Knoll, Florian and Johnson, Patricia},
  booktitle={International conference on medical image computing and computer-assisted intervention},
  pages={64--73},
  year={2020},
  organization={Springer}
}

@inproceedings{zhang2018multi,
  title={Multi-channel generative adversarial network for parallel magnetic resonance image reconstruction in k-space},
  author={Zhang, Pengyue and Wang, Fusheng and Xu, Wei and Li, Yu},
  booktitle={International conference on medical image computing and computer-assisted intervention},
  pages={180--188},
  year={2018},
  organization={Springer}
}

@article{guo2021reference,
  title={Reference-based magnetic resonance image reconstruction using texture transformer},
  author={Guo, Pengfei and Patel, Vishal M},
  journal={arXiv preprint arXiv:2111.09492},
  year={2021}
}

@article{guo2023reconformer,
  title={Reconformer: Accelerated mri reconstruction using recurrent transformer},
  author={Guo, Pengfei and Mei, Yiqun and Zhou, Jinyuan and Jiang, Shanshan and Patel, Vishal M},
  journal={IEEE transactions on medical imaging},
  volume={43},
  number={1},
  pages={582--593},
  year={2023},
  publisher={IEEE}
}

@inproceedings{ronneberger2015u,
  title={U-net: Convolutional networks for biomedical image segmentation},
  author={Ronneberger, Olaf and Fischer, Philipp and Brox, Thomas},
  booktitle={International Conference on Medical image computing and computer-assisted intervention},
  pages={234--241},
  year={2015},
  organization={Springer}
}

@inproceedings{li2026sd,
  title={SD-FSMIS: Adapting Stable Diffusion for Few-Shot Medical Image Segmentation},
  author={Li, Meihua and Zhang, Yang and He, Weizhao and Qu, Hu and Li, Yisong},
  booktitle={Proceedings of the IEEE/CVF Conference on Computer Vision and Pattern Recognition},
  pages={29979--29989},
  year={2026}
}

@inproceedings{huix2024natural,
  title={Are natural domain foundation models useful for medical image classification?},
  author={Huix, Joana Pal{\'e}s and Ganeshan, Adithya Raju and Haslum, Johan Fredin and S{\"o}derberg, Magnus and Matsoukas, Christos and Smith, Kevin},
  booktitle={Proceedings of the IEEE/CVF winter conference on applications of computer vision},
  pages={7634--7643},
  year={2024}
}

@inproceedings{hashmi2025can,
  title={Can Natural Domain Foundation Models Be Applied to Cardiac MRI Reconstruction?},
  author={Hashmi, Anam and Dietlmeier, Julia and Maniparambil, Mayug and Garcia-Cabrera, Carles and Curran, Kathleen M and O'Connor, Noel},
  booktitle={Medical Imaging with Deep Learning-Short Papers},
  year={2025}
}

@inproceedings{zhu2024melo,
  title={Melo: Low-rank adaptation is better than fine-tuning for medical image diagnosis},
  author={Zhu, Yitao and Shen, Zhenrong and Zhao, Zihao and Wang, Sheng and Wang, Xin and Zhao, Xiangyu and Shen, Dinggang and Wang, Qian},
  booktitle={2024 IEEE International Symposium on Biomedical Imaging (ISBI)},
  pages={1--5},
  year={2024},
  organization={IEEE}
}

@inproceedings{hatamizadeh2022unetr,
  title={Unetr: Transformers for 3d medical image segmentation},
  author={Hatamizadeh, Ali and Tang, Yucheng and Nath, Vishwesh and Yang, Dong and Myronenko, Andriy and Landman, Bennett and Roth, Holger R and Xu, Daguang},
  booktitle={Proceedings of the IEEE/CVF winter conference on applications of computer vision},
  pages={574--584},
  year={2022}
}

@inproceedings{radford2021learning,
  title={Learning transferable visual models from natural language supervision},
  author={Radford, Alec and Kim, Jong Wook and Hallacy, Chris and Ramesh, Aditya and Goh, Gabriel and Agarwal, Sandhini and Sastry, Girish and Askell, Amanda and Mishkin, Pamela and Clark, Jack and others},
  booktitle={International conference on machine learning},
  pages={8748--8763},
  year={2021},
  organization={PmLR}
}

@article{zhang2023biomedclip,
  title={Biomedclip: a multimodal biomedical foundation model pretrained from fifteen million scientific image-text pairs},
  author={Zhang, Sheng and Xu, Yanbo and Usuyama, Naoto and Xu, Hanwen and Bagga, Jaspreet and Tinn, Robert and Preston, Sam and Rao, Rajesh and Wei, Mu and Valluri, Naveen and others},
  journal={arXiv preprint arXiv:2303.00915},
  year={2023}
}

@article{oquab2023dinov2,
  title={Dinov2: Learning robust visual features without supervision},
  author={Oquab, Maxime and Darcet, Timoth{\'e}e and Moutakanni, Th{\'e}o and Vo, Huy and Szafraniec, Marc and Khalidov, Vasil and Fernandez, Pierre and Haziza, Daniel and Massa, Francisco and El-Nouby, Alaaeldin and others},
  journal={arXiv preprint arXiv:2304.07193},
  year={2023}
}

@article{larkman2007parallel,
  title={Parallel magnetic resonance imaging},
  author={Larkman, David J and Nunes, Rita G},
  journal={Physics in Medicine \& Biology},
  volume={52},
  number={7},
  pages={R15--R55},
  year={2007}
}

@article{pruessmann2006encoding,
  title={Encoding and reconstruction in parallel MRI},
  author={Pruessmann, Klaas P},
  journal={NMR in Biomedicine: An International Journal Devoted to the Development and Application of Magnetic Resonance In vivo},
  volume={19},
  number={3},
  pages={288--299},
  year={2006},
  publisher={Wiley Online Library}
}

@inproceedings{wang2016accelerating,
  title={Accelerating magnetic resonance imaging via deep learning},
  author={Wang, Shanshan and Su, Zhenghang and Ying, Leslie and Peng, Xi and Zhu, Shun and Liang, Feng and Feng, Dagan and Liang, Dong},
  booktitle={2016 IEEE 13th international symposium on biomedical imaging (ISBI)},
  pages={514--517},
  year={2016},
  organization={IEEE}
}

@article{liang2020deep,
  title={Deep magnetic resonance image reconstruction: Inverse problems meet neural networks},
  author={Liang, Dong and Cheng, Jing and Ke, Ziwen and Ying, Leslie},
  journal={IEEE Signal Processing Magazine},
  volume={37},
  number={1},
  pages={141--151},
  year={2020},
  publisher={IEEE}
}

@inproceedings{kumar2025medvisionllama,
  title={Medvisionllama: Leveraging pre-trained large language model layers to enhance medical image segmentation},
  author={Kumar, Gurucharan Marthi Krishna and Chadha, Aman and Mendola, Janine D and Shmuel, Amir},
  booktitle={Proceedings of the IEEE/CVF International Conference on Computer Vision},
  pages={1114--1124},
  year={2025}
}

@article{anand2023one,
  title={One-shot localization and segmentation of medical images with foundation models},
  author={Anand, Deepa and Singhal, Vanika and Shanbhag, Dattesh D and KS, Shriram and Patil, Uday and Bhushan, Chitresh and Manickam, Kavitha and Gui, Dawei and Mullick, Rakesh and Gopal, Avinash and others},
  journal={arXiv preprint arXiv:2310.18642},
  year={2023}
}

@article{baharoon2023evaluating,
  title={Evaluating general purpose vision foundation models for medical image analysis: An experimental study of dinov2 on radiology benchmarks},
  author={Baharoon, Mohammed and Qureshi, Waseem and Ouyang, Jiahong and Xu, Yanwu and Aljouie, Abdulrhman and Peng, Wei},
  journal={arXiv preprint arXiv:2312.02366},
  year={2023}
}

@inproceedings{muller2022joint,
  title={Joint learning of localized representations from medical images and reports},
  author={M{\"u}ller, Philip and Kaissis, Georgios and Zou, Congyu and Rueckert, Daniel},
  booktitle={European conference on computer vision},
  pages={685--701},
  year={2022},
  organization={Springer}
}

@inproceedings{song2024dino,
  title={Dino-reg: General purpose image encoder for training-free multi-modal deformable medical image registration},
  author={Song, Xinrui and Xu, Xuanang and Yan, Pingkun},
  booktitle={International Conference on Medical Image Computing and Computer-Assisted Intervention},
  pages={608--617},
  year={2024},
  organization={Springer}
}

@article{lian2024less,
  title={Less could be better: Parameter-efficient fine-tuning advances medical vision foundation models},
  author={Lian, Chenyu and Zhou, Hong-Yu and Yu, Yizhou and Wang, Liansheng},
  journal={arXiv preprint arXiv:2401.12215},
  year={2024}
}

@article{dutt2023parameter,
  title={Parameter-efficient fine-tuning for medical image analysis: The missed opportunity},
  author={Dutt, Raman and Ericsson, Linus and Sanchez, Pedro and Tsaftaris, Sotirios A and Hospedales, Timothy},
  journal={arXiv preprint arXiv:2305.08252},
  year={2023}
}

@article{hu2022lora,
  title={Lora: Low-rank adaptation of large language models.},
  author={Hu, Edward J and Shen, Yelong and Wallis, Phillip and Allen-Zhu, Zeyuan and Li, Yuanzhi and Wang, Shean and Wang, Liang and Chen, Weizhu and others},
  journal={Iclr},
  volume={1},
  number={2},
  pages={3},
  year={2022}
}

@inproceedings{cekmeceli2024vision,
  title={Do vision foundation models enhance domain generalization in medical image segmentation?},
  author={Cekmeceli, Kerem and Himmetoglu, Meva and Tombak, Guney I and Susmelj, Anna and Erdil, Ertunc and Konukoglu, Ender},
  booktitle={European Conference on Computer Vision},
  pages={185--200},
  year={2024},
  organization={Springer}
}

@article{vaswani2017attention,
  title={Attention is all you need},
  author={Vaswani, Ashish and Shazeer, Noam and Parmar, Niki and Uszkoreit, Jakob and Jones, Llion and Gomez, Aidan N and Kaiser, {\L}ukasz and Polosukhin, Illia},
  journal={Advances in neural information processing systems},
  volume={30},
  year={2017}
}

@article{dosovitskiy2020image,
  title={An image is worth 16x16 words: Transformers for image recognition at scale},
  author={Dosovitskiy, Alexey and Beyer, Lucas and Kolesnikov, Alexander and Weissenborn, Dirk and Zhai, Xiaohua and Unterthiner, Thomas and Dehghani, Mostafa and Minderer, Matthias and Heigold, Georg and Gelly, Sylvain and others},
  journal={arXiv preprint arXiv:2010.11929},
  year={2020}
}

@article{ba2016layer,
  title={Layer normalization},
  author={Ba, Jimmy Lei and Kiros, Jamie Ryan and Hinton, Geoffrey E},
  journal={arXiv preprint arXiv:1607.06450},
  year={2016}
}

@inproceedings{cai2024vip,
  title={Vip-llava: Making large multimodal models understand arbitrary visual prompts},
  author={Cai, Mu and Liu, Haotian and Mustikovela, Siva Karthik and Meyer, Gregory P and Chai, Yuning and Park, Dennis and Lee, Yong Jae},
  booktitle={Proceedings of the IEEE/CVF Conference on Computer Vision and Pattern Recognition},
  pages={12914--12923},
  year={2024}
}

@inproceedings{zanella2024low,
  title={Low-rank few-shot adaptation of vision-language models},
  author={Zanella, Maxime and Ben Ayed, Ismail},
  booktitle={Proceedings of the IEEE/CVF Conference on Computer Vision and Pattern Recognition},
  pages={1593--1603},
  year={2024}
}

@article{wang2023cmrxrecon,
  title={CMRxRecon: an open cardiac MRI dataset for the competition of accelerated image reconstruction},
  author={Wang, Chengyan and Lyu, Jun and Wang, Shuo and Qin, Chen and Guo, Kunyuan and Zhang, Xinyu and Yu, Xiaotong and Li, Yan and Wang, Fanwen and Jin, Jianhua and others},
  journal={arXiv preprint arXiv:2309.10836},
  year={2023}
}

@article{larsson2003snr,
  title={SNR-optimality of sum-of-squares reconstruction for phased-array magnetic resonance imaging},
  author={Larsson, Erik G and Erdogmus, Deniz and Yan, Rui and Principe, Jose C and Fitzsimmons, Jeffrey R},
  journal={Journal of Magnetic Resonance},
  volume={163},
  number={1},
  pages={121--123},
  year={2003},
  publisher={Elsevier}
}

@inproceedings{xiang2023diffcmr,
  title={Diffcmr: fast cardiac mri reconstruction with diffusion probabilistic models},
  author={Xiang, Tianqi and Yue, Wenjun and Lin, Yiqun and Yang, Jiewen and Wang, Zhenkun and Li, Xiaomeng},
  booktitle={International Workshop on Statistical Atlases and Computational Models of the Heart},
  pages={380--389},
  year={2023},
  organization={Springer}
}

@article{wang2025cmrxrecon2024,
  title={CMRxRecon2024: a multimodality, multiview k-space dataset boosting universal machine learning for accelerated cardiac MRI},
  author={Wang, Zi and Wang, Fanwen and Qin, Chen and Lyu, Jun and Ouyang, Cheng and Wang, Shuo and Li, Yan and Yu, Mengyao and Zhang, Haoyu and Guo, Kunyuan and others},
  journal={Radiology: Artificial Intelligence},
  volume={7},
  number={2},
  pages={e240443},
  year={2025},
  publisher={Radiological Society of North America}
}

@inproceedings{wang2003multiscale,
  title={Multiscale structural similarity for image quality assessment},
  author={Wang, Zhou and Simoncelli, Eero P and Bovik, Alan C},
  booktitle={The thrity-seventh asilomar conference on signals, systems \& computers, 2003},
  volume={2},
  pages={1398--1402},
  year={2003},
  organization={Ieee}
}

\section{Supplementary Material}
\label{app:appendix}
\subsection{Additional LoRA Rank Ablations}

We provide the complete LoRA rank sensitivity results for CLIP, BiomedCLIP, and DINOv2 across CMRxRecon2023 and CMRxRecon2024. Unless otherwise stated, the optimal LoRA rank was selected based on overall reconstruction performance across SSIM, PSNR, and NMSE. Preference was given to configurations achieving the best results in the majority of metrics. In cases where no single rank consistently dominated across all three metrics, SSIM was prioritized, as it better reflects perceptual image quality and structural fidelity in MRI reconstruction, with PSNR and NMSE used as secondary criteria.

\paragraph{BiomedCLIP Rank Sensitivity.}
Table~\ref{tab:app_biomedclip_rank} presents the effect of the LoRA rank on BiomedCLIP. Increasing the rank increases the number of trainable parameters from 34.0M ($r=8$) to 38.4M ($r=128$), providing greater adaptation capacity. Across both CMRxRecon2023 and CMRxRecon2024, performance generally improves with rank, with $r=128$ consistently achieving the strongest overall trade-off between SSIM, PSNR, and NMSE. The gains are particularly noticeable on CMRxRecon2024, suggesting that larger adaptation capacity enables more effective specialization of the pretrained biomedical representations. Consequently, $r=128$ was selected for all BiomedCLIP experiments reported in the main paper.

\begin{table*}[htbp]
\centering
\scriptsize
\setlength{\tabcolsep}{4pt}
\renewcommand{\arraystretch}{1.08}
\caption{LoRA rank ablation for BiomedCLIP on CMRxRecon2023 and CMRxRecon2024. Trainable parameter counts are reported in millions (M). Trainable parameter counts include the LoRA modules, feature fusion layers, and reconstruction decoder. The best-performing rank for each acceleration factor is highlighted in bold.}
\label{tab:app_biomedclip_rank}

\begin{tabular}{c|c|c|ccc|ccc}
\toprule
\multirow{2}{*}{Dataset} &
\multirow{2}{*}{Rank} &
\multirow{2}{*}{Params (M)}
& \multicolumn{3}{c|}{$4\times$}
& \multicolumn{3}{c}{$8\times$} \\
\cmidrule(lr){4-6}
\cmidrule(lr){7-9}
&
&
& SSIM$\uparrow$ & PSNR$\uparrow$ & NMSE$\downarrow$
& SSIM$\uparrow$ & PSNR$\uparrow$ & NMSE$\downarrow$ \\
\midrule

\multirow{3}{*}{CMRxRecon2023}
& 8   & 34.0 & 0.8973 & 31.8867 & 0.0356 & 0.8679 & 30.3138 & \textbf{0.0508} \\
& 64  & 36.1 & \textbf{0.8975} & 31.8523 & 0.0375 & 0.8672 & 29.9981 & 0.0586 \\
& \textbf{128} & 38.4 & 0.8972 & \textbf{31.9133} & \textbf{0.0351} & \textbf{0.8706} & \textbf{30.4793} & 0.0516 \\

\midrule

\multirow{3}{*}{CMRxRecon2024}
& 8   & 34.0 & 0.8917 & 31.6805 & 0.0416 & 0.8484 & 29.5407 & 0.0707 \\
& 64  & 36.1 & 0.8925 & 31.8170 & 0.0409 & 0.8509 & 29.8524 & 0.0640 \\
& \textbf{128} & 38.4 & \textbf{0.8945} & \textbf{31.8295} & \textbf{0.0407} & \textbf{0.8543} & \textbf{29.9158} & \textbf{0.0636} \\

\bottomrule
\end{tabular}

\vspace{0.2cm}

\begin{tabular}{c|c|c|ccc}
\toprule
\multirow{2}{*}{Dataset} &
\multirow{2}{*}{Rank} &
\multirow{2}{*}{Params (M)}
& \multicolumn{3}{c}{$10\times$} \\
\cmidrule(lr){4-6}
&
&
& SSIM$\uparrow$ & PSNR$\uparrow$ & NMSE$\downarrow$ \\
\midrule

\multirow{3}{*}{CMRxRecon2023}
& 8   & 34.0 & 0.8573 & 29.5979 & 0.0602 \\
& 64  & 36.1 & 0.8567 & \textbf{29.6908} & \textbf{0.0593} \\
& \textbf{128} & 38.4 & \textbf{0.8605} & 29.6505 & 0.0605 \\

\midrule

\multirow{3}{*}{CMRxRecon2024}
& 8   & 34.0 & 0.8387 & 29.1110 & 0.0758 \\
& 64  & 36.1 & 0.8434 & 29.3223 & 0.0710 \\
& \textbf{128} & 38.4 & \textbf{0.8459} & \textbf{29.3922} & \textbf{0.0709} \\

\bottomrule
\end{tabular}
\end{table*}

\paragraph{CLIP Rank Sensitivity.}
Table~\ref{tab:app_clip_rank} presents the LoRA rank ablation for CLIP. Increasing the rank raises the number of trainable parameters from 34.1M ($r=8$) to 40.8M ($r=128$), providing greater adaptation capacity. Unlike BiomedCLIP, no single rank consistently achieves the best performance across all datasets and acceleration factors. Higher ranks generally perform best at lower acceleration factors, with $r=128$ achieving the strongest results at $4\times$ acceleration on both datasets and at $8\times$ acceleration on CMRxRecon2023. However, for more challenging reconstruction settings, intermediate-rank adaptation often provides a better trade-off. In particular, $r=64$ achieves the best overall performance at $10\times$ acceleration on both datasets and at $8\times$ acceleration on CMRxRecon2024. Based on these results, $r=128$ was selected for CLIP at $4\times$ and $8\times$ acceleration on CMRxRecon2023 and at $4\times$ acceleration on CMRxRecon2024, while $r=64$ was used for the remaining settings reported in the main paper.

\begin{table*}[htbp]
\centering
\scriptsize
\setlength{\tabcolsep}{4pt}
\renewcommand{\arraystretch}{1.08}
\caption{LoRA rank ablation for CLIP on CMRxRecon2023 and CMRxRecon2024. Trainable parameter counts are reported in millions (M). Trainable parameters include the LoRA modules, feature fusion layers, and reconstruction decoder. The best-performing rank for each acceleration factor is highlighted in bold.}
\label{tab:app_clip_rank}

\begin{tabular}{c|c|c|ccc|ccc}
\toprule
\multirow{2}{*}{Dataset} &
\multirow{2}{*}{Rank} &
\multirow{2}{*}{Params (M)}
& \multicolumn{3}{c|}{$4\times$}
& \multicolumn{3}{c}{$8\times$} \\
\cmidrule(lr){4-6}
\cmidrule(lr){7-9}
&
&
& SSIM$\uparrow$ & PSNR$\uparrow$ & NMSE$\downarrow$
& SSIM$\uparrow$ & PSNR$\uparrow$ & NMSE$\downarrow$ \\
\midrule

\multirow{3}{*}{CMRxRecon2023}
& 8   & 34.1 & \textbf{0.9020} & 32.0410 & 0.0351 & 0.8699 & \textbf{30.5080} & 0.0498 \\
& 64  & 37.2 & 0.9009 & 32.0592 & 0.0344 & 0.8703 & 30.2384 & 0.0530 \\
& \textbf{128} & 40.8 & 0.9013 & \textbf{32.1458} & \textbf{0.0340} & \textbf{0.8710} & 30.4831 & \textbf{0.0492} \\

\midrule

\multirow{3}{*}{CMRxRecon2024}
& 8   & 34.1 & 0.8967 & 31.9873 & 0.0394 & 0.8571 & 29.9309 & 0.0645 \\
& \textbf{64}(for $8\times$)  & 37.2 & 0.8992 & 32.1980 & \textbf{0.0374} & 0.8591 & \textbf{30.2022} & \textbf{0.0585} \\
& \textbf{128}(for $4\times$) & 40.8 & \textbf{0.9003} & \textbf{32.2644} & 0.0377 & \textbf{0.8592} & 30.0607 & 0.0613 \\

\bottomrule
\end{tabular}

\vspace{0.2cm}

\begin{tabular}{c|c|c|ccc}
\toprule
\multirow{2}{*}{Dataset} &
\multirow{2}{*}{Rank} &
\multirow{2}{*}{Params (M)}
& \multicolumn{3}{c}{$10\times$} \\
\cmidrule(lr){4-6}
&
&
& SSIM$\uparrow$ & PSNR$\uparrow$ & NMSE$\downarrow$ \\
\midrule

\multirow{3}{*}{CMRxRecon2023}
& 8   & 34.1 & 0.8620 & 29.8363 & 0.0558 \\
& \textbf{64} & 37.2 & \textbf{0.8629} & \textbf{29.9360} & 0.0549 \\
& 128 & 40.8 & 0.8619 & 29.8862 & \textbf{0.0548} \\

\midrule

\multirow{3}{*}{CMRxRecon2024}
& 8   & 34.1 & 0.8478 & 29.3569 & 0.0712 \\
& \textbf{64} & 37.2 & 0.8500 & \textbf{29.5561} & \textbf{0.0675} \\
& 128 & 40.8 & \textbf{0.8506} & 29.4911 & 0.0695 \\

\bottomrule
\end{tabular}
\end{table*}

\paragraph{DINOv2 Rank Sensitivity.}
Table~\ref{tab:app_dinov2_rank} presents the LoRA rank ablation for DINOv2. Increasing the adaptation rank increases the number of trainable parameters from 34.1M ($r=8$) to 40.8M ($r=128$). Overall, DINOv2 exhibits the strongest preference for higher-rank adaptation among the evaluated foundation models. On CMRxRecon2024, performance consistently improves with increasing rank across all acceleration factors, with $r=128$ achieving the best reconstruction quality in terms of SSIM, PSNR, and NMSE. Similar trends are observed on CMRxRecon2023 at $4\times$ and $8\times$ acceleration. The only exception occurs at $10\times$ acceleration on CMRxRecon2023, where an intermediate rank of $r=64$ provides the best balance between reconstruction fidelity and error reduction. These results suggest that DINOv2 benefits substantially from increased adaptation capacity, likely due to the richness of its self-supervised visual representations. Consequently, $r=128$ was selected for all DINOv2 experiments except for the $10\times$ CMRxRecon2023 setting, where $r=64$ was used.

\begin{table*}[htbp]
\centering
\scriptsize
\setlength{\tabcolsep}{4pt}
\renewcommand{\arraystretch}{1.08}
\caption{LoRA rank ablation for DINOv2 on CMRxRecon2023 and CMRxRecon2024. Trainable parameter counts are reported in millions (M). Trainable parameters include the LoRA modules, feature fusion layers, and reconstruction decoder. The best-performing rank for each acceleration factor is highlighted in bold.}
\label{tab:app_dinov2_rank}
\begin{tabular}{c|c|c|ccc|ccc}
\toprule
\multirow{2}{*}{Dataset} &
\multirow{2}{*}{Rank} &
\multirow{2}{*}{Params (M)}
& \multicolumn{3}{c|}{$4\times$}
& \multicolumn{3}{c}{$8\times$} \\
\cmidrule(lr){4-6}
\cmidrule(lr){7-9}
&
&
& SSIM$\uparrow$ & PSNR$\uparrow$ & NMSE$\downarrow$
& SSIM$\uparrow$ & PSNR$\uparrow$ & NMSE$\downarrow$ \\
\midrule

\multirow{3}{*}{CMRxRecon2023}
& 8   & 34.1 & 0.9050 & \textbf{32.4577} & \textbf{0.0316} & 0.8768 & 30.3741 & 0.0524 \\
& 64  & 37.2 & 0.9058 & 32.2943 & 0.0324 & 0.8758 & 30.7677 & 0.0452 \\
& \textbf{128} & 40.8 & \textbf{0.9061} & 32.4175 & 0.0319 & \textbf{0.8776} & \textbf{30.9313} & \textbf{0.0441} \\

\midrule

\multirow{3}{*}{CMRxRecon2024}
& 8   & 34.1 & 0.9023 & 32.2638 & 0.0365 & 0.8628 & 30.0431 & 0.0623 \\
& 64  & 37.2 & 0.9025 & 32.2664 & 0.0379 & 0.8636 & 30.0201 & 0.0632 \\
& \textbf{128} & 40.8 & \textbf{0.9054} & \textbf{32.5112} & \textbf{0.0353} & \textbf{0.8663} & \textbf{30.5859} & \textbf{0.0543} \\

\bottomrule
\end{tabular}

\vspace{0.2cm}

\begin{tabular}{c|c|c|ccc}
\toprule
\multirow{2}{*}{Dataset} &
\multirow{2}{*}{Rank} &
\multirow{2}{*}{Params (M)}
& \multicolumn{3}{c}{$10\times$} \\
\cmidrule(lr){4-6}
&
&
& SSIM$\uparrow$ & PSNR$\uparrow$ & NMSE$\downarrow$ \\
\midrule

\multirow{3}{*}{CMRxRecon2023}
& 8   & 34.1 & 0.8643 & 30.0642 & 0.0535 \\
& \textbf{64} & 37.2 & 0.8649 & \textbf{30.1798} & \textbf{0.0511} \\
& 128 & 40.8 & \textbf{0.8667} & 29.8821 & 0.0586 \\

\midrule

\multirow{3}{*}{CMRxRecon2024}
& 8   & 34.1 & 0.8533 & 29.7096 & 0.0651 \\
& 64  & 37.2 & 0.8544 & \textbf{29.9021} & \textbf{0.0622} \\
& \textbf{128} & 40.8 & \textbf{0.8571} & 29.8925 & 0.0635 \\

\bottomrule
\end{tabular}
\end{table*}

\end{document}